\documentclass[letterpaper, 10 pt, conference]{ieeeconf}  

\usepackage{graphicx}
\usepackage{flushend}
\usepackage{balance}
\usepackage{xcolor}
\usepackage{multirow}
\usepackage{multicol}
\usepackage{url}

\IEEEoverridecommandlockouts                              

\title{\LARGE \bf
Developing Modular Grasping and Manipulation Pipeline Infrastructure to Streamline Performance Benchmarking
}

\author{Brian Flynn$^{1}$, Kostas Bekris$^{2}$, Berk Calli$^{3}$, Aaron Dollar$^{4}$, Adam Norton$^{1}$, Yu Sun$^{5}$, and Holly Yanco$^{1}$
\thanks{*This work was supported in part by the National Science Foundation (awards TI-2346069 and CNS-1925604) and the National Institute of Standards and Technology (awards 60NANB24D294 and 70NANB20H199).}
\thanks{$^{1}$New England Robotics Validation and Experimentation (NERVE) Center, University of Massachusetts Lowell, Lowell, MA, USA
        {\tt\small brian\_flynn,adam\_norton,holly\_yanco@uml.edu}}%
\thanks{$^{2}$Rutgers University, New Brunswick, NJ, USA
        {\tt\small kostas.bekris@cs.rutgers.edu}}%
\thanks{$^{3}$Worcester Polytechnic University, Worcester, MA, USA
        {\tt\small bcalli@wpi.edu}}%
\thanks{$^{4}$Yale University, New Haven, CT, USA
        {\tt\small aaron.dollar@yale.edu}}%
\thanks{$^{5}$University of South Florida, Tampa, FL, USA
        {\tt\small yusun@usf.edu}}%
}

\begin{document}

\maketitle
\thispagestyle{empty}
\pagestyle{empty}




\section{Introduction and Background}

Based on our work engaging with the robot manipulation community through online surveys, holding conference workshops, and deep dive interviews with community stakeholders, we have determined several limitations of the current open-source and benchmarking ecosystem with recommendations for improvement.
Current limitations include issues integrating open-source products and a lack of truly modular software to enable component-level and holistic system evaluations.
This aligns with previously published sentiments around issues with code reproducibility~\cite{cervera2023run,cervera2018try} and defining benchmarks that evaluate components of a manipulation solution, such as the Cluttered Environment Picking Benchmark~\cite{d2023cluttered}.
Recommendations for improvement include organized repositories of open-source products and benchmarking assets, and the benefit of modular software components was highlighted.
Projects like SceneReplica~\cite{khargonkar2024scenereplica} demonstrate the effectiveness of benchmarking with interchangeable pipeline components (perception, grasp planning, motion planning, and control), and the GRASPA project~\cite{bottarel2023graspa} provides an example implementation on the Franka Panda robot to benchmark the performance of three grasp planners.

With these factors in mind, we are developing a modular grasping and manipulation pipeline infrastructure for improved performance benchmarking.
As part of the COMPARE Ecosystem project~\cite{compare-web}, we intend to use this infrastructure to establish standards and guidelines towards improving open-source development and benchmarking.
This paper presents a brief review of experiments conducted during development of the pipeline infrastructure, the advancements enabled by each experiment, an overview of the current architecture of the pipeline infrastructure, and future work.

\section{Pipeline Infrastructure}

The goal of the pipeline infrastructure is to dictate the flow of an experiment using state machines with nested behavior trees, wherein users will utilize existing ROS packages or create interchangeable components (e.g., grasp planning, motion planning, perception) following established guidelines to ``drop in.''
Components should be interchangeable to be readily swapped out without having to write new code (e.g., each grasp planning service utilizes the same request and response message type such as image in, pose out) and the user should make minimal modifications to the infrastructure set-up to match their testing needs.
The infrastructure is also intended to be sufficiently hardware-agnostic (i.e., functional regardless of hardware, assuming that the hardware is ROS compatible and a ROS driver exists for such hardware).

It is likely that ROS1 is still the preferred implementation for many research groups given that ROS2 is still largely in development, so streamlined simulation environment integrations and certain robot drivers may not yet exist.
The community is expected to shift to ROS2 at large due to the fact that ROS1 is nearing end-of-life, so we are converting the codebase to ROS2 despite the fact that the pipeline infrastructure was initially developed and tested using ROS1 assets.
Various components of the pipeline infrastructure are in development and can be found on our organization Github page\footnote{https://github.com/uml-robotics} (e.g., components for point cloud processing\footnote{https://github.com/uml-robotics/robot-common-3d}).

Our prior pipeline infrastructure development in ROS1 was built around FlexBe\footnote{http://philserver.bplaced.net/fbe/} as a state-machine-driven manipulation process.
It used concise states that make requests to ROS services and actionservers to execute certain tasks (e.g., motion planning via MoveIt, filtering point clouds via PCL) that could be easily added, removed, or reordered as needed using the FlexBe user interface or directly through code; these states are wrapped into a process container called a behavior. 
The manipulation utilities we developed\footnote{https://github.com/uml-robotics/armada\_behaviors} were designed to be robot-agnostic and typically relied on the ROS parameter server to handle varying hardware requirements.

While FlexBe is undergoing development for ROS2 environments\footnote{https://github.com/FlexBE/flexbe\_behavior\_engine}, it was built upon a python state-machine library called SMACH\footnote{http://wiki.ros.org/smach} which is not inherently dependent on ROS1 or ROS2. 
The SMACH library can be used to provide nearly identical functionality albeit without the aid of the FlexBe UI which, while helpful, is not a necessary component. 
This allows us to continue development on the pipeline infrastructure in ROS2 without relying on the completion of certain external resources.

\section{Experimentation}

\begin{figure}
\includegraphics[width=\columnwidth]{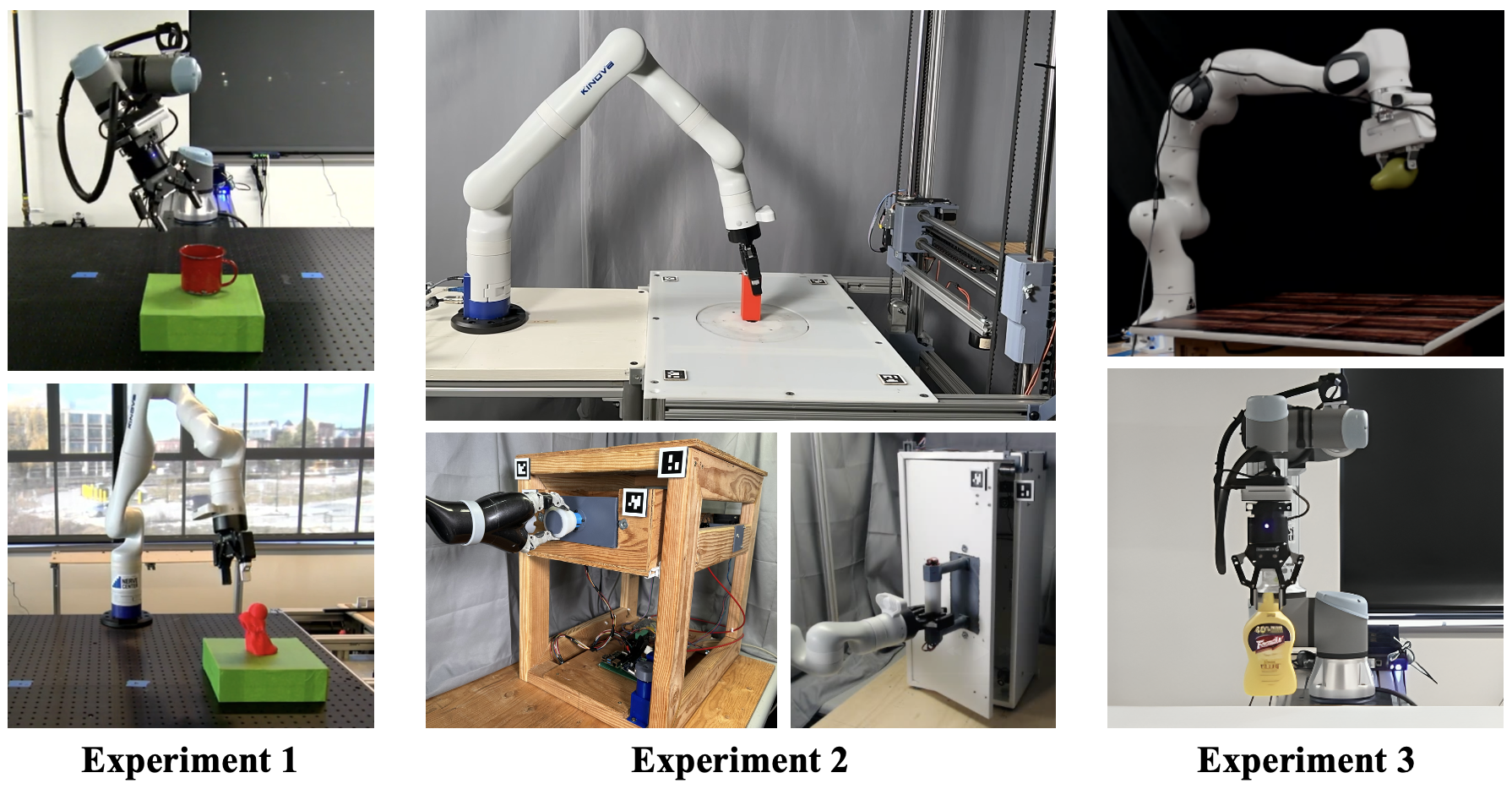}
\caption{Experiments performed during development.} 
\vspace{-0.5cm}
\label{fig:experiments}
\end{figure}

During development of the pipeline infrastructure, multiple experiments were conducted to exercise the infrastructure in our labs at UML and WPI as well as our partners' labs at OSU  (see Fig.~\ref{fig:experiments}).
All of these experiments led to further improvements to the modularity of the infrastructure to allow for more variety of components that could be easily implemented within the pipeline. 
They also validated the functionality of the pipeline to work with code contributed by multiple internal and external users.

\textbf{Experiment 1}: Performance of the Grasp Pose Detection (GPD)~\cite{ten2017grasp} algorithm was evaluated by deploying it within the initial FlexBe manipulation pipeline infrastructure across two robot embodiments (UR5e and Kinova Gen3 with the same gripper and sensor) at UML.
Following the grasp planning benchmarking protocol in Bekiroglu et al.~\cite{bekiroglu2019benchmarking}, 640 grasps were performed on two objects under two lighting conditions and with or without 10 cm workspace elevation (unpublished).
Each individual robot, gripper, and sensor embodiment was controlled using the same FlexBe behavior with the only change being a set of parameters which handled lower-level differences between systems.


\textbf{Experiment 2}: State machine functionality was expanded to control three physical test apparatuses that automatically reset task conditions (object repositioning after grasping, door and drawer closing after opening) to enable automated, repeated testing for multiple trials. 
The manipulation behavior was nested inside of a test apparatus reset behavior which controlled data recording and automated resetting of apparatuses between manipulation operations.
Using a Kinova Gen3, our partners at OSU performed 1,020 grasps on four objects using the grasp reset mechanism~\cite{graspreset}, and 600 grasps using a Kinova Jaco2 were performed on the door and drawer reset mechanisms~\cite{doordrawerreset}. 


\textbf{Experiment 3}: The modularity of the pipeline was further expanded by integrating four vision-based grasping algorithms (GG-CNN, ResNet, Top Surface, and Mask-Based).
These algorithms were originally used in a pipeline developed at WPI on a Franka Emika Panda with the Panda gripper. 
The grasping algorithms were compatible with the FlexBe pipeline architecture because they used ROS service interactions; requesting a sensor image and returning a pose or list of poses, allowing them to effectively swap places with the grasping algorithm initially used within the infrastructure.
Using ten objects against textured or non-textured backgrounds, two lighting conditions, and again following the grasp planning benchmarking protocol in Bekiroglu et al.~\cite{bekiroglu2019benchmarking}, 5,040 grasps were performed and their performance compared~\cite{vision-based-benchmarking}.
A Docker set-up to reproduce this benchmarking is available on GitHub\footnote{https://github.com/vinayakkapoor/vision\_based\_grasping\_benchmarking}.
\textcolor{white}{\footnote{https://github.com/moveit/moveit\_grasps}}



\section{Future Work}

\begin{table}[]
\begin{tabular}{|p{1.7cm}|p{1.5cm}|p{4.2cm}|}
\hline
\textbf{\begin{tabular}[c]{@{}l@{}}Grasp\\ Planner\end{tabular}} & \textbf{Input Data} & \textbf{Output Pose} \\ \hline
\multirow{2}{*}{\begin{tabular}[c]{@{}l@{}}Dex-Net\\ (GQ-CNN)~\cite{mahler2017dex}\end{tabular}} & \multirow{2}{*}{Depth image} & 2D grasp rectangle (x, y, width,   height, angle) \\ \hline
\multirow{3}{*}{GQ-CNN~\cite{mahler2019learning}} & \multirow{3}{*}{Depth image} & 2D grasp rectangle (x, y, width, height, angle) with grasp success probability \\ \hline
GPD~\cite{ten2017grasp} & Point cloud & 6-DoF grasp pose (x, y, z, r, p, y) \\ \hline
PointNetGPD
\cite{liang2019pointnetgpd} & \multirow{2}{*}{Point cloud} & \multirow{2}{*}{6-DoF grasp pose (x, y, z, r, p, y)} \\ \hline
GraspNet~\cite{mousavian2019graspnet} & Point cloud & 6-DoF grasp pose (x, y, z, r, p, y) \\ \hline
Contact-GraspNet~\cite{sundermeyer2021contact} & \multirow{2}{*}{Point cloud} & \multirow{2}{*}{6-DoF grasp pose (x, y, z, r, p, y)} \\ \hline
\multirow{2}{*}{GraspIt!~\cite{miller2004graspit}} & 3D object model mesh & 6-DoF grasp pose (x, y, z, r, p, y) with force closure metrics \\ \hline
\multirow{3}{*}{MoveIt Grasps$^8$} & Collision object in planning scene & Heuristic grasp poses based on object position; 6-DoF grasp pose (x, y, z, r, p, y) \\ \hline
\end{tabular}
\caption{Grasp planners to be integrated.}
\vspace{-1cm}
\label{tab:grasp-planners}
\end{table}


The pipeline infrastructure continues to be developed to expand its modularity by adding more grasp planners, improved instructions to easily change or add custom motion planners, and integrate perception components.
In order to ease implementation of new components, we will analyze the required inputs and outputs of each type of component towards the development of standards and guidelines for modularization.
For example, Table~\ref{tab:grasp-planners} shows several grasp planners that we intend to integrate with the particular input data each utilizes to plan grasps and the format of their output poses.
Based on this type of analysis, a set of classes can be defined for components with similar requirements.
This will be a topic of discussion and development for the community working together through the COMPARE Ecosystem project~\cite{compare-web}.
Similar exercises will be conducted for motion planners (e.g., OMPL~\cite{sucan2012the-open-motion-planning-library}, CHOMP~\cite{ratliff2009chomp}, STOMP~\cite{kalakrishnan2011stomp}, Pilz\footnote{https://wiki.ros.org/pilz\_industrial\_motion}, TrajOpt~\cite{schulman2013finding}; MoveIt also provides tutorials for creating custom plugins that can be leveraged) and perception (e.g., object segmentation, object recognition, pose estimation).

Additional experiments are planned to exercise the pipeline functionality across five collaborating labs (UML, WPI, Yale, Rutgers, and USF) each with different robot hardware and varying levels of researcher experience.
As part of the COMPARE project, a seasonal school hackathon event is being organized for summer 2025 where the pipeline will be used as a baseline for continued development and deployment of additional components, enabling significant comparison benchmarking to be conducted.

\bibliographystyle{ieeetr}
\bibliography{references}

\end{document}